\documentclass[10pt,twocolumn,letterpaper]{article}

\usepackage{cvpr}

\definecolor{cvprblue}{rgb}{0.21,0.49,0.74}
\usepackage[pagebackref,breaklinks,colorlinks,allcolors=cvprblue]{hyperref}
\usepackage[misc]{ifsym}
\def\paperID{} 
\def\confName{CVPR}
\def\confYear{2026}

\title{Addressing Exacerbated Attention Sink for\\ Source-Free Cross-Domain Few-Shot Learning}

\author{Shuai Yi, Yixiong Zou\textsuperscript{\Letter}, Yuhua Li\textsuperscript{\Letter}, Ruixuan Li\textsuperscript{\Letter}\\ Huazhong University of Science and Technology\\
	{\tt\small \{yishuai, yixiongz, idcliyuhua, rxli\}@hust.edu.cn}}

\begin{document}
\maketitle
\begin{abstract}
Vision-language models (VLMs) like CLIP have shown impressive generalization capabilities, yet their potential for Cross-Domain Few-Shot Learning (CDFSL) remains underexplored, where the model needs to transfer source-domain information to target domains with scarce training data. 
While the attention sink phenomenon has been observed in VLMs for certain tasks, its role in CDFSL scenarios has not been studied. 
In this paper, we uncover a critical issue overlooked by prior works: standard target-domain few-shot fine-tuning in CDFSL significantly exacerbates the attention sink problem, leading to poor discriminability across classes. 
To understand this phenomenon, through extensive experiments, we interpret it as the model's shortcut learning for domain adaptation: to overcome the huge domain gap between the source and target domains, the model shows a high tendency to push tokens that are initially closer to target-domain classes (i.e., simple tokens) to be even closer to these classes, exacerbating the attention sink and wasting the capability of learning other discriminative but initially further tokens (i.e., hard tokens).
To address this, we propose a novel approach to dynamically re-weight tokens according to their relevance with target-domain classes during the target-domain finetuning, which explicitly suppresses the model's reliance on these simple tokens and enhances the learning of hard tokens, reducing sink tokens and enhancing discriminability. 
Extensive experiments on four benchmark datasets validate the rationale of our method, demonstrating new state-of-the-art performance. Our codes are available at https://github.com/shuaiyi308/TIR.
\end{abstract}    
\section{Introduction}
\label{sec:intro}
\vspace{-0.1cm}
Cross-Domain Few-Shot Learning (CDFSL)~\cite{zou2024flatten,yi2025revisiting} seeks to adapt models pre-trained on large-scale source-domain datasets (e.g., ImageNet) to specialized target domains (e.g., medical or satellite imagery) where only very few labeled examples are accessible. As large pre-trained models become more widespread, source-free fine-tuning, where no source-domain data is available during target-domain fine-tuning, has emerged as a more practical task~\cite{xu2024enhancing} in CDFSL. Although vision-language models (VLM)~\cite{zhao2023clip} such as CLIP have shown strong generalization across a variety of vision tasks, their potential for few-shot fine-tuning in cross-domain scenarios remains relatively underexplored.

\begin{figure}[t]
	\centering
	\includegraphics[width=1.0\columnwidth]{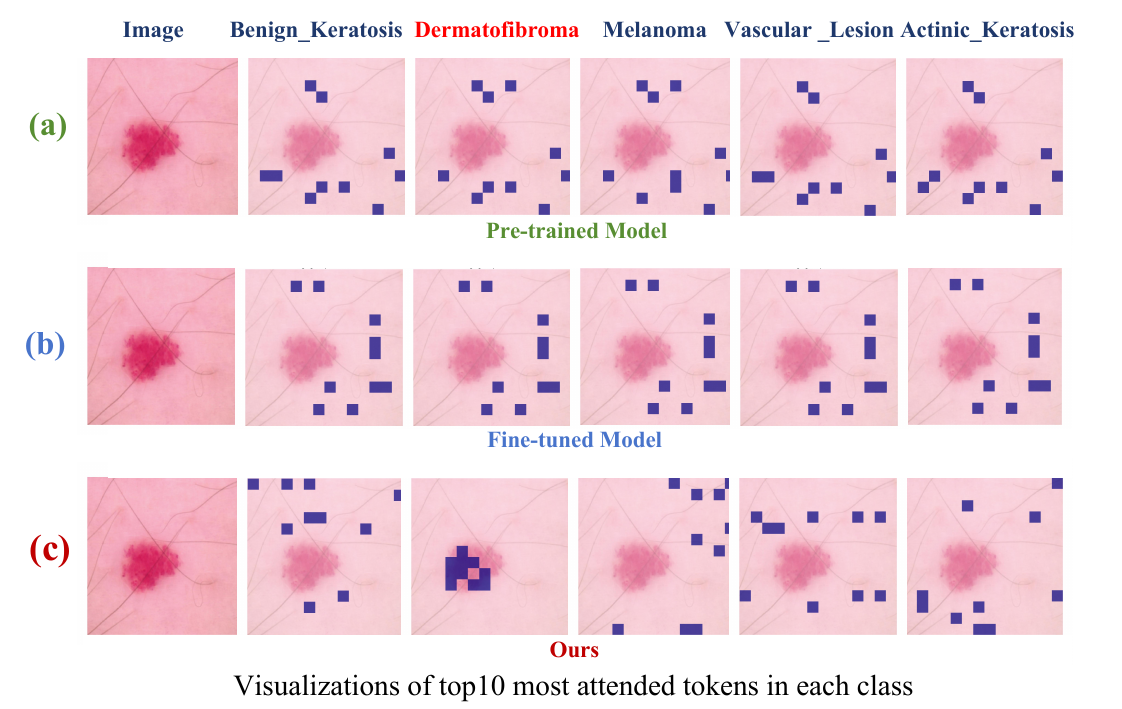}
    \vspace{-0.7cm}
	\caption{
    (a) We visualize highly attended visual tokens for different target-domain classes of the pre-trained VLM (before fine-tuning), finding that although substantial overlap exists across classes (i.e., attention sink), minor differences still remain. 
    (b) However, after standard target-domain few-shot fine-tuning, these highly attended tokens become completely identical across different classes, demonstrating significant exacerbation of the attention sink phenomenon and harming the discriminability. 
    This reveals a critical question: why does the few-shot fine-tuning intensify the attention sink problem in VLMs? In this paper, we will delve into this phenomenon for an interpretation, and propose methods based on it for better target-domain fine-tuning.
    (c) With our method, the attention sink is significantly alleviated, where different visual tokens now properly attend to their corresponding semantic concepts, greatly improving the discriminability.}\vspace{-0.3cm}
	\label{fig:sink}
\end{figure}


In VLMs, attention sink, as a widely observed phenomenon~\cite{gu2024attention}, is a double-edged sword to help the model aggregate information~\cite{xiao2023efficient}, but also leads to hallucination~\cite{wang2025mirage,zhang2024seeing}. However, in this paper, we find a rarely studied phenomenon: \textbf{the few-shot fine-tuning process in CDFSL significantly exacerbates the attention sink phenomenon}, leading to only the negative side of it. For example, as shown in \cref{fig:sink}, we visualize visual tokens that are the most attended by different class texts, finding that before fine-tuning, although these high-attention tokens exhibit substantial overlap across classes (\cref{fig:sink}a), some minor difference in the attention still exists; however, after cross-domain few-shot fine-tuning, they become totally identical (\cref{fig:sink}b). 
This phenomenon raises a critical question: why does cross-domain few-shot fine-tuning significantly intensify the attention sink problem?

In this paper, we delve into this phenomenon for an interpretation and a solution. 
Through extensive experiments, we find that these exacerbated sink tokens contain excessive domain information, and are much closer to \textbf{all} target-domain classes \textbf{even without fine-tuning}. This indicates that due to the huge domain gap between the source and target domains, the model pays excessive attention to those \textbf{simple tokens} that are initially easier to align with target-domain classes for easier domain adaptation, therefore absorbing excessive domain information. \textit{This tendency of fine-tuning acts as a \textbf{shortcut} that reduces the learning of discriminative tokens} (i.e., attended by only a specific class, which is measured to be much further than those simple tokens and is thus \textbf{harder} to learn).
In a word, such a shortcut of domain adaptation significantly exacerbates the attention sink on these simple tokens.

In contrast, in the source domain, since it is similar to VLMs' pre-training data, we find that although the attention sink exists, the model can still majorly focus on tokens attended by a specific class. 
Therefore, we need to shrink the model's tendency from learning those simple tokens (i.e., sink tokens, inherently closer to target-domain classes) to those harder tokens (i.e., discriminative tokens, further than simple tokens to target-domain classes), just like the token distribution on the source domain.

Based on the above interpretation, to achieve this goal, we design a simple yet effective method to dynamically re-weight the token according to its relevance with the fine-tuning classes, which smoothly assigns simple tokens with lower weights and those discriminative but harder ones with higher weights.
Extensive experiments on four CDFSL benchmarks with large domain gaps demonstrate that we successfully shift the learning of simple (sink) tokens to the learning of harder (discriminative) tokens, thereby significantly reducing sink tokens (\cref{fig:sink}c) and outperforming state-of-the-art approaches.

In summary, our contributions can be listed as follows.
\begin{itemize}
\item To the best of our knowledge, we are the first to find that the target-domain fine-tuning in CDFSL significantly amplifies the attention sink.

\item We delve into this phenomenon for an interpretation, finding the model's shortcut in learning those simple tokens that are initially closer to all target-domain classes for easier domain adaptation, sacrificing the learning of harder (further) but more discriminative tokens, leading to the significantly amplified sink problem.

\item Based on this interpretation, we further propose a method to explicitly suppress the model's shortcut in learning of simple tokens, shifting its learning from simple tokens to those harder but more discriminative ones, thereby addressing the sink problem and improving performance.

\item Extensive experiments on four benchmark datasets validate our rationale and new state-of-the-art performance.
\end{itemize}

\vspace{-0.1cm}
\section{Delve into Exacerbated Attention Sinks}
\label{sec:analysis and interpretation}
\vspace{-0.1cm}

\subsection{Preliminaries}
\label{sec:preliminaries}
\textbf{Source-Free Cross-Domain Few-Shot Learning:}
 Given a target domain dataset ${D}_T$, an episode $E = \{(S, Q), Y\}$ is randomly sampled for meta-testing, formulated as an N-way K-shot problem\cite{zou2024flatten,xu2023deep}. Specifically, for each episode $E$ from ${D}_T$, $N$ classes with $K$ labeled images are sampled to form the support set $S$, and the same $N$ classes with $M$ different images comprise the query set $Q$. The label set for these $N$ classes is denoted as $Y = \{{c_i}\}_{i=1}^N$. The support set $S$ is utilized for training the model, while the query set $S$ is used to evaluate accuracy.

\vspace{0.1cm}
\noindent\textbf{Fine-tuning in CLIP:} For classification, a prompt template (e.g., ``a photo of a \{dog\}") is used to generate textual descriptions for each class\cite{zhang2025decoupling}. Let $\mathbf{r}_k$ be the tokenized prompt for class $k$, the text encoder produces its embedding $\mathbf{t}_k = F_t(\mathbf{r}_k)$. Similarly, each image $\mathbf{x}_i$ is processed by the visual encoder $F_v$ to obtain the normalized visual embedding  $\mathbf{f}_i = F_v(\mathbf{x}_i)$. The cross-entropy loss for the task defined over the image-text similarity scores is:
\begin{equation}
\setlength{\abovedisplayskip}{5pt}
\setlength{\belowdisplayskip}{5pt}
\mathcal{L}_{\text{cross}} = -\frac{1}{N}\sum_i \log \frac{\exp(\text{sim}(\mathbf{f}_i, \mathbf{t}_i)/\tau)}{\sum_j \exp(\text{sim}(\mathbf{f}_i, \mathbf{t}_j)/\tau)},
\label{eq:cross_entropy}
\end{equation}
where $\text{sim}(\mathbf{f}_i, \mathbf{t}_j) = \frac{\mathbf{f}_i^\intercal \mathbf{t}_j}{|\mathbf{f}_i||\mathbf{t}_j|}$ denotes cosine similarity between image sample $i$ and text prompt $j$, and $\tau$ is a temperature coefficient, which is typically set to 0.01 in CLIP.

\begin{figure}[t]
	\centering
	\includegraphics[width=1.02\columnwidth]{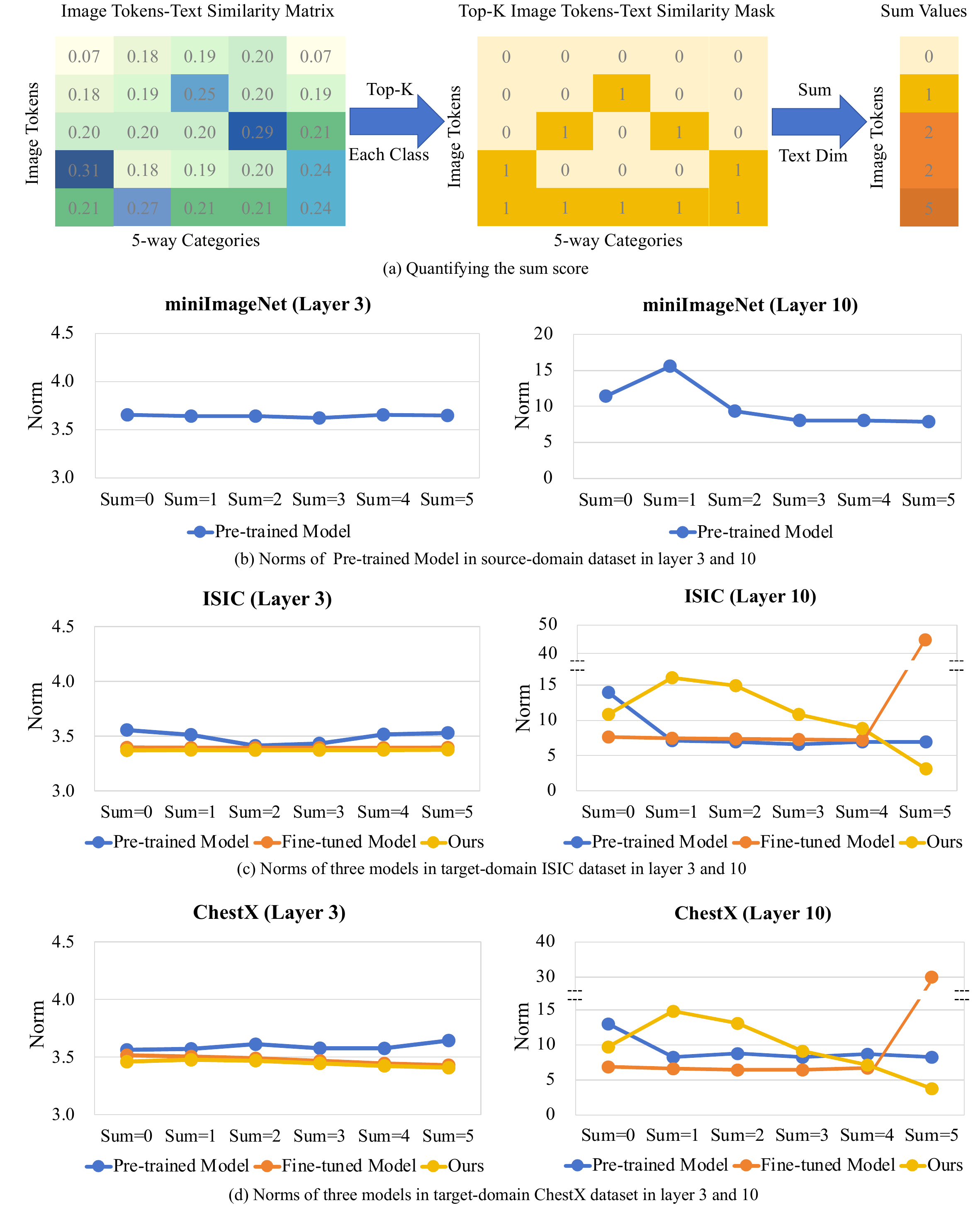}
    \vspace{-0.7cm}
	\caption{
    (a) Illustration of the sum score calculation: we compute the similarity between each image token and all class text embeddings, then categorize tokens by their cross-class activation patterns, where ``Sum=5" indicates tokens active across all classes while ``Sum=1" represents class-specific discriminative tokens. 
    (b) At layer 3, both source and target domain models show similar attention distributions with weak semantic awareness. At layer 10, the source domain model develops specialized attention to discriminative tokens (``Sum=1"), while the target domain fine-tuned model shifts toward non-discriminative tokens (``Sum=5") with significantly higher norms, demonstrating that fine-tuning, instead of merely cross-domain transfer, serves as the primary driver of attention sink exacerbation.
    }\vspace{-0.3cm}
	\label{fig:layer_norm`}
\end{figure}

\subsection{Target-domain fine-tuning exacerbates attention sink}
Given our finding that cross-domain fine-tuning exacerbates the visual attention sink, we then investigate whether this phenomenon originates primarily from the cross-domain shift or the fine-tuning process itself. Following previous work\cite{wu2024attention}, we employ the norm value to identify the sink tokens and conduct a layer-wise analysis of semantic awareness on the 5-way 5-shot task.

Specifically, we compute the similarity between each image token and all class name embeddings, then analyze tokens with top-30\% cosine similarities with each class. As shown in Fig.~\ref{fig:layer_norm`}a, if the token shows the top-30\% similarity to a given class, then the corresponding box is set to 1. By summing up the number of classes that a token shows top-30\% similarity to, we can identify the degree to which this token is attended by all classes.
Therefore, ``Sum=5" represents tokens highly active across all five classes, meaning it loses the discriminability to classes, while "Sum=1" indicates discriminative tokens unique to a single class (see Fig.~\ref{fig:method} for methodological details).

\vspace{0.1cm}
\noindent\textbf{Cross-Domain Impact.} 
As shown in \cref{fig:layer_norm`}b, we measure the norm of the ``Sum=1,2,..,5'' tokens, representing the importance that the model assigns to these tokens~\cite{gu2024attention}.
In the source domain, the token norm of different sum numbers is similar in shallow layers (e.g., layer 3, more layers are in the appendix), showing weak semantic awareness in shallow layers. 
With the layer going deeper, the model progressively develops specialized norms to discriminative tokens (``Sum=1") in deeper layers (layer 10), indicating the discriminability appears in deep layers. 
However, although similar norm distribution is in the shallow layers of target domains (the pre-training line on ISIC~\cite{codella2019skin} and ChestX~\cite{Wang_2017}, more datasets are in the appendix), the specialized norm on discriminative tokens (``Sum=1'') does not appear, demonstrating the fundamental challenge of cross-domain shift lies in the failed transfer of complex semantic features, leaving target-domain models with limited effective patterns to reuse for target-domain fine-tuning.

\vspace{0.1cm}
\noindent\textbf{Fine-Tuning Impact.} The fine-tuning process reveals a different but more critical pattern. While shallow layers remain mostly similar to the pre-training line, we observe significantly increased norm of ``Sum=5'' tokens in deep layers. 
This behavior fundamentally differs from that of the source domain\footnote{Since the source-domain data is similar to the pre-training data of CLIP, we assume the pre-trained model on the source-domain dataset does not need to be further fine-tuned}, where the model focuses on the most discriminative image tokens(``Sum=1"). 
Moreover, since the token norm is significantly higher than most tokens, it is indeed the sink token. In other words, the standard target-domain fine-tuning causes the exacerbation of attention sinks.

\begin{figure}[t]
	\centering
	\includegraphics[width=0.9\columnwidth]{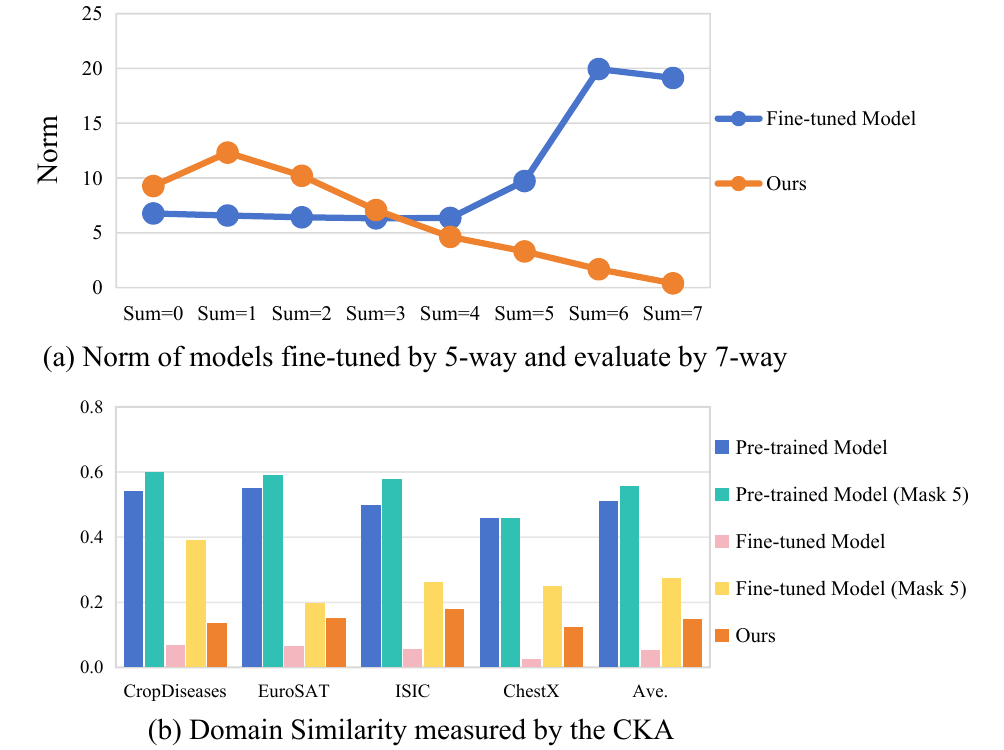}
    \vspace{-0.2cm}
	\caption{ (a) In the 7-way evaluation of models fine-tuned on 5-way classes, the norm distribution demonstrates that the model primarily focuses on tokens showing high similarity to all classes in the domain, including classes outside the training classes. This indicates that fine-tuning drives the model to learn domain-level information, instead of merely the given training classes. (b) To verify this, we use the CKA similarity to measure the domain similarity, where a higher similarity indicates less domain information. Results validate that ``Sum=5'' tokens capture domain information: removing these tokens increases the domain similarity.
}\vspace{-0.3cm}
	\label{fig:cka}
\end{figure}

\subsection{Sink tokens absorb domain information}

While fine-tuning indeed improves performance, this raises a fundamental question: why does the fine-tuning process tend to learn sink tokens? 
To answer this question, we need to know what information is contained in these sink tokens.
Since these tokens are fine-tuned to be close to all fine-tuned target-domain classes, we first test whether these tokens are also fine-tuned to be close to target-domain classes that are not fine-tuned on.
Therefore, we fine-tune the model in a 5-way 5-shot setting but evaluate it on a 7-way test setup. 
As shown in \cref{fig:cka}a, results reveal that the model increasingly attends to tokens showing high similarity across more categories, including those outside the training set, rather than focusing solely the five training classes. This indicates that fine-tuning drives the model to adapt to the entire domain, rather than just the current 5-way task.

Therefore, we further hypothesize that these sink tokens mainly contain domain information that is crucial for target-domain adaptation.
To verify it, we follow \cite{zou2024compositional,zoucloser,ma2024reconstruction} to use Centered Kernel Alignment (CKA) similarity to measure the domain similarity between the source and target domains, where higher domain similarity means lower domain information. 
We test four models: the pretrained model with/without masking ``Sum=5'' tokens, and the fine-tuned model with/without masking these tokens (denote the masking of ``Sum=5'' tokens as mask5 in \cref{fig:cka}b).
As shown in \cref{fig:cka}b, we have three findings:

(1) Removing ``Sum=5'' tokens of the pre-trained model increases the CKA value, indicating these tokens inherently carry domain information; 

(2) After fine-tuning, the CKA value drops significantly, confirming the model absorbs target-domain information; 

(3) Removing ``Sum=5'' tokens after fine-tuning greatly elevates CKA, demonstrating they capture substantial domain information.

Together, these results validate that ``Sum=5'' tokens serve as a role to absorb target-domain information, enabling the model to adapt to target domains.

\subsection{Shortcut of domain adaptation exacerbates attention sink}

Based on the above experiments, we understand that the model tends to adapt to the target domain against huge domain gaps during fine-tuning, but why are these tokens selected (instead of those discriminative ``Sum=1'' tokens)?

\begin{figure}[t]
	\centering
	\includegraphics[width=1.0\columnwidth]{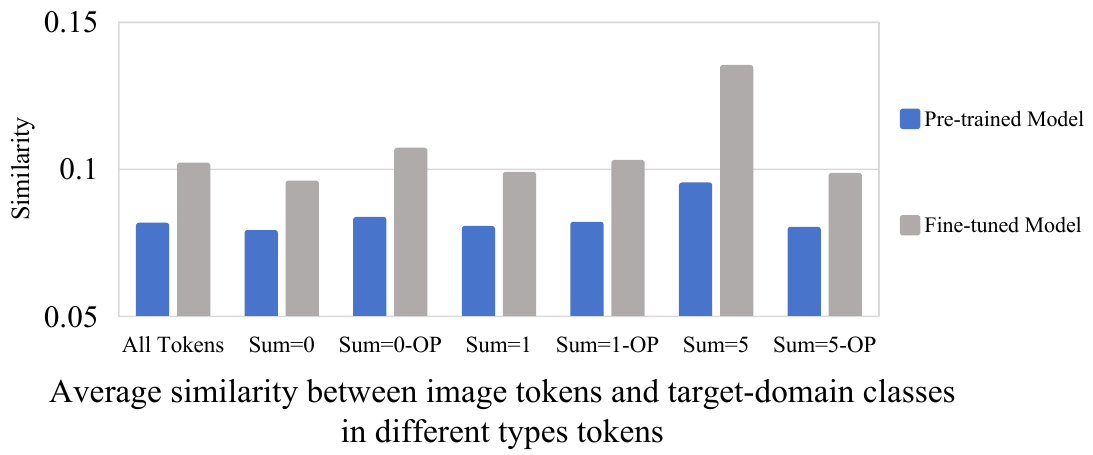}
    \vspace{-0.7cm}
	\caption{Quantitative analysis reveals that ``Sum=5'' tokens exhibit higher visual-text similarity than others even without fine-tuning, showing these tokens are easier to be learned to align with target-domain classes for domain adaptation. Therefore, we conclude that the exacerbated attention sink problem is essentially a shortcut of domain adaptation, sacrificing the learning of other discriminative tokens (``Sum=1'') whose visual-text similarity is smaller and are thus harder to align with target-domain classes.
    }\vspace{-0.2cm}
	\label{fig:distance}
\end{figure}

Therefore, we quantitatively analyze the properties of these tokens. We measure the average similarity between different tokens and target-domain fine-tuning classes, including ``Sum=0'', ``Sum=1'', ``Sum=5'' tokens, and tokens other than ``Sum=0'', ``Sum=1'', ``Sum=5'' tokens, denoted as ``Sum=0\_OP'', ``Sum=1\_OP'', ``Sum=5\_OP'' tokens, respectively.
As in \cref{fig:distance}, we find that: 

(1) Those ``Sum=5'' tokens exhibit higher similarity compared to other tokens, even without fine-tuning; 

(2) The cross-modal similarity consistently increases, and that of ``Sum=5'' tokens increases the most.


Since \cite{xu2020cross} pointed out the cross-modal misalignment problem under cross-domain scenarios, one behavior of domain adaptation is to re-align the visual and text features, represented as increasing the cross-modal similarity. Therefore, we draw the following conclusion:

Under huge domain gaps between the source and target domains, the pre-trained model can hardly fit the target domain. The fine-tuning process drives the model to learn to adapt to the target domain (verified in \cref{fig:cka}), represented as re-aligning the visual and text features (verified in \cref{fig:distance}). However, the model can reuse only a limited set of pre-trained patterns (verified in \cref{fig:layer_norm`}), and only scarce data is available for training. Therefore, it is difficult for the model to learn useful patterns during fine-tuning. \textbf{As a shortcut, the model seeks to align those simple visual tokens (``Sum=5'' tokens) that are initially closer to class texts} (without fine-tuning, validated in \cref{fig:distance}), \textbf{which are easier to align with these texts for domain adaptation}. As a result, excessive domain information is stored in these tokens, leading to significantly increased norms and exacerbating the attention sink.

However, such a shortcut learning process sacrifices the learning of discriminative tokens (``Sum=1''), as the initial visual-text similarity of these tokens is smaller than ``Sum=5'' tokens. Therefore, we need to redirect the learning of ``Sum=5'' tokens to other discriminative tokens, which we aim to achieve in the next section.

\begin{figure*}[t]
\vspace{-0.4cm}
	\centering
    
	\includegraphics[width=2.0\columnwidth]{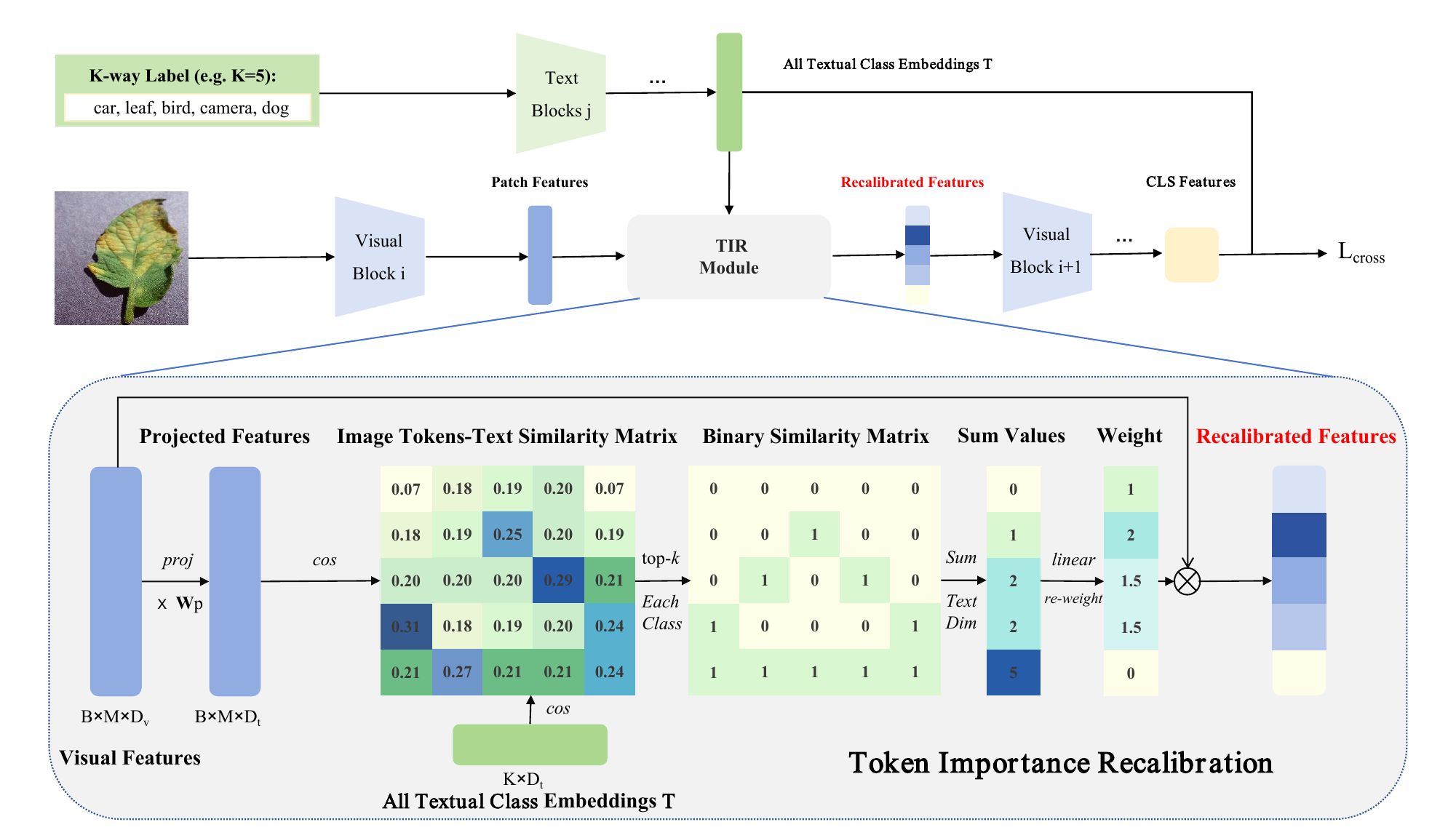}
    \vspace{-0.3cm}
  \caption{Overview of our Token Importance Recalibration (TIR) framework. The image and text inputs are processed through the CLIP encoders to obtain patch features and textual embeddings.  Between specific transformer blocks (blocks 8-9 and 10-11), we compute the cosine similarity between projected visual tokens and all class text embeddings, then identify sink tokens and discriminative tokens through top-k selection. Based on the cross-class activation pattern, we compute sum score and apply our linear re-weighting scheme to suppress attention sink tokens (``$\text{sum}=5$'') while enhancing discriminative tokens (``$\text{sum}=1$''). The recalibrated features are then propagated through subsequent layers, enabling the model to focus on those hard tokens for improved cross-domain classification.}
  \vspace{-0.3cm}
  \label{fig:method}
\end{figure*}

\vspace{-0.2cm}
\section{Method}
\vspace{-0.1cm}


Based on the above analysis and interpretation, we develop a simple yet effective Token Importance Recalibration (TIR) approach that dynamically re-weights visual tokens according to their semantic discriminability during target-domain fine-tuning. Our method is inserted between specific transformer blocks in the CLIP visual encoder (specifically in deeper layers), where we empirically find the attention sink phenomenon becomes most pronounced.

As illustrated in Fig.\ref{fig:method},  given a batch of input images $\mathbf{I} \in \mathbb{R}^{B \times H \times W \times C}$, they are first processed through the initial transformer blocks to obtain visual features $\mathbf{V} \in \mathbb{R}^{B \times M \times D_v}$, where $B$ denotes batch size, $M$ the number of image tokens, and $D_v$ the visual feature dimension. These features are extracted from specific intermediate blocks (typically after the 8th and 10th blocks). We first project these visual features to align with the text embedding space:
\begin{equation}
\setlength{\abovedisplayskip}{5pt}
\setlength{\belowdisplayskip}{5pt}
\mathbf{V}' = \text{LayerNorm}(\mathbf{V}) \mathbf{W}_p
\label{eq:v2t}
\end{equation}
where $\mathbf{W}_p \in \mathbb{R}^{D_v \times D_t}$ is the visual-text projection matrix in CLIP and $D_t$ represents the text feature dimension.

We then compute the cosine similarity between each projected visual token and all textual class embeddings $\mathbf{T} \in \mathbb{R}^{K \times D_t}$, where $K$ is the number of classes (typically $K=5$ in 5-way setting):
\begin{equation}
\setlength{\abovedisplayskip}{5pt}
\setlength{\belowdisplayskip}{5pt}
s_{b,i,j} = \frac{\mathbf{v}'_{b,i} \cdot \mathbf{t}_j}{\|\mathbf{v}'_{b,i}\| \|\mathbf{t}_j\|}, \quad b \in [1,B], i \in [1,M], j \in [1,K]
\label{eq:cos_sim}
\end{equation}
To identify tokens with high similarity to specific classes, we construct a binary similarity matrix by selecting the top-$k$ most similar tokens for each class:
\begin{equation}
\setlength{\abovedisplayskip}{5pt}
\setlength{\belowdisplayskip}{5pt}
\mathbf{S}^{\text{binary}}_{b,i,j} = 
\begin{cases} 
1 & \text{if } s_{b,i,j} \in \text{top-}k(\mathbf{s}_{b,:,j}) \\
0 & \text{otherwise}
\end{cases}
\label{eq:binary}
\end{equation}
The Sum score, that can identify the degree to which this token is attended by all classes, is computed by aggregating its binary indicators across all classes:
\begin{equation}
\setlength{\abovedisplayskip}{5pt}
\setlength{\belowdisplayskip}{5pt}
\text{Sum}_{b,i} = \sum_{j=1}^{K} \mathbf{S}^{\text{binary}}_{b,i,j}
\end{equation}
Based on the empirical observation that tokens with $\text{Sum}_{b,i} = 5$ exhibit attention sink behavior while tokens with $\text{Sum}_{b,i} = 1$ represent class-discriminative patterns, we apply a conditional linear re-weighting scheme:
\begin{equation}
\setlength{\abovedisplayskip}{5pt}
\setlength{\belowdisplayskip}{5pt}
w_{b,i} = = 
\begin{cases} 
 1 - \beta \times (\text{Sum}_{b,i} - \alpha) & \text{if  } \text{Sum} > 0  \\
1 & \text{otherwise}
\end{cases}
\label{eq:sum}
\end{equation}
where $\alpha = 3$ serves as the neutrality threshold and $\beta = 0.5$ controls the suppression/enhancement intensity. This formulation suppresses non-discriminative sink tokens ($\text{Sum}_{b,i} = 5 \rightarrow w_{b,i} = 0$) while enhancing discriminative ones ($\text{Sum}_{b,i} = 1 \rightarrow w_{b,i} = 2$).

\begin{table*}[t]
\vspace{-0.3cm}
	\begin{center}
     \caption{Comparison with state-of-the-art works by the 5-way 1-shot and 5-way 5-shot classification.}\vspace{-0.3cm}
      \label{tab:sotas}
		\resizebox{1.0\textwidth}{!}{
			\begin{tabular}{lcccccccccc}
				\toprule
				Method & Mark &Backbone & Shot & Source & Target & ISIC & EuroSAT & CropDiseases  & ChestX & Ave. \\
				\midrule
				StyleAdv-FT~\cite{fu2023styleadv} & CVPR-23 &ViT/DINO &  1 & $\checkmark$& $\checkmark$ & 33.99 & 74.93 & 84.11 & 22.92& 53.99 \\
				FLoR~\cite{zou2024flatten} &CVPR-24  &ViT/DINO  & 1 & $\checkmark$ &$\checkmark$  & 35.49 & 73.09 & 83.55 & 23.26& 53.85 \\
                DAMIM~\cite{ma2024reconstruction} &AAAI-25 &ViT/DINO    & 1 & $\checkmark$ & $\checkmark$  & 36.35 & 73.61 & 83.90& 23.38 & 54.31
                \\
                CD-CLS~\cite{zoucloser} & NeurIPS-24&ViT/DINO  & 1 & $\checkmark$ & $\checkmark$ & 35.56 & 74.97 & 84.53 &  23.39 & 54.62
                \\
               AttnTemp~\cite{zouattention} & NeurIPS-24 &ViT/DINO &  1 & $\checkmark$ &$\checkmark$ & 38.05 & 75.09 & 84.78 & 23.63 & 55.39
                \\
				ReCIT~\cite{yi2025revisiting} &ICML-25 &ViT/DINO & 1 & $\checkmark$ &  $\checkmark$  & 38.48 & 75.23 & 85.92 & 23.84& 55.87\\
				REAP~\cite{yi2025random} &ICML-25 &ViT/DINO & 1 & $\checkmark$ &  $\checkmark$  & 38.67 & 75.97 & 85.33 & \textbf{24.17}& 56.04\\
                FN+VDB~\cite{yazdanpanah2022visual} &CVPR-22 &RN18 & 1 &  - &  $\checkmark$ & 32.96 & 69.67 & 79.68 & 22.64 & 51.24\\
                IM-DCL~\cite{xu2024enhancing} &TIP-24 &RN10 & 1 &  - &  $\checkmark$  & 38.13 & 77.14 & 84.37 & 23.98& 55.91\\
                StepSTP~\cite{xu2024step} &TPAMI-25 &ViT/CLIP & 1 &  - &  $\checkmark$  & 32.97 & 70.01 & 84.84 & 22.84& 52.68\\
                CLIP-LoRA-Vision~\cite{zanella2024low} &CVPRW-24 &ViT/CLIP & 1 &  - &  $\checkmark$  & 35.23 & 81.41 & 85.32  & 21.73 & 55.92 \\
                \textbf{CLIP-LoRA-Vision + TIR}&Ours&ViT/CLIP & 1 & - &  $\checkmark$& \textbf{39.38}& \textbf{82.53} & \textbf{86.91}  & \underline{23.98}& \textbf{58.20  }\\
                \midrule
				StyleAdv-FT~\cite{fu2023styleadv} & CVPR-23 &ViT/DINO    & 5 & $\checkmark$ &  $\checkmark$  & 51.23 & 90.12 & 95.99 &26.97& 66.08 \\
				FLoR~\cite{zou2024flatten} &CVPR-24  &ViT/DINO  & 5 & $\checkmark$ &$\checkmark$ & 53.06 & 90.75 & 96.47 & 27.02& 66.83 \\
                DAMIM~\cite{ma2024reconstruction} &AAAI-25 &ViT/DINO    & 5 & $\checkmark$ & $\checkmark$  & 54.86 & 91.18 & 96.34 & 27.82& 67.55
                \\
                CD-CLS~\cite{zoucloser} & NeurIPS-24&ViT/DINO  & 5 & $\checkmark$ & $\checkmark$  & 54.69 & 91.53 & 96.27 &27.66& 67.54
                \\
                AttnTemp~\cite{zouattention} & NeurIPS-24&ViT/DINO  & 5 & $\checkmark$ & $\checkmark$  & 54.91 & 90.82 & 96.66 &28.03& 67.61
                \\
				ReCIT~\cite{yi2025revisiting} &ICML-25 &ViT/DINO & 5 & $\checkmark$ &  $\checkmark$ & 54.91 & 91.58 &  96.85& 28.88 & 68.06 \\
				REAP~\cite{yi2025random} &ICML-25 &ViT/DINO & 5 & $\checkmark$ &  $\checkmark$  & 55.28 & 91.79 &  96.71& 28.34& 68.03 \\
                FN+VDB~\cite{yazdanpanah2022visual} &CVPR-22 &RN18 & 5 &  - &  $\checkmark$  & 47.48 & 87.31 & 94.63& 25.55 & 64.74\\
                IM-DCL~\cite{xu2024enhancing} &TIP-24 &RN10 & 5 &  - &  $\checkmark$  & 52.74 & 89.47 & 95.73 & \textbf{28.93}& 66.72\\
                StepSTP~\cite{xu2024step} &TPAMI-25 &ViT/CLIP & 5 &  - &  $\checkmark$  & 52.12 & 89.40 & 96.01 & 26.36& 65.97\\
                CLIP-LoRA-Vision~\cite{zanella2024low} &CVPRW-24  &ViT/CLIP & 5 &  - &  $\checkmark$  & 51.10 & 92.52 & 96.21 & 24.13& 65.99  \\
                \textbf{CLIP-LoRA-Vision + TIR}&Ours &ViT/CLIP & 5 & - &  $\checkmark$  & \textbf{56.73} & \textbf{93.49} & \textbf{97.42} & 26.12& \textbf{68.44 } \\
				\bottomrule
        \end{tabular}}
	\end{center}\vspace{-0.6cm}
   
\end{table*}

The recalibrated visual features are computed as:
\begin{equation}
\setlength{\abovedisplayskip}{5pt}
\setlength{\belowdisplayskip}{5pt}
\mathbf{V}^{\text{weighted}} = \mathbf{w} \odot \mathbf{V}
\label{eq:reweight}
\end{equation}
where $\odot$ denotes element-wise multiplication and the weight tensor $\mathbf{w} \in \mathbb{R}^{B \times M}$ is broadcast appropriately.
The recalibrated features are then passed through the remaining transformer blocks. The final class embedding $\mathbf{V}_{\text{class}}$ is obtained from the last layer (typically the [CLS] token) and compared with textual embeddings to compute the cross-entropy loss as illustrated in Eq.~\ref{eq:cross_entropy}

This approach enables the model to focus on semantically discriminative patterns while mitigating the attention sink effect. This approach effectively shifts the model's learning focus from sink tokens to class-discriminative tokens, improving target-domain fine-tuning.

As further analyzed in the appendix, this insight yields a simplified strategy that is effective in practical $K$-way $N$-shot settings: setting the weight of Sum=$K$ tokens to $0$ while assigning Sum=$1$ tokens any weight greater than $1$ achieves nearly equivalent performance with minimal intervention. This simplified strategy maintains the performance benefits of our complete method while offering greater practicality and easier implementation for various few-shot learning configurations.

\vspace{-0.1cm}
\section{Experiments}
\label{sec:exp}
\vspace{-0.1cm}

\subsection{Experimental Settings}
\textbf{Datasets.} Based on established benchmarks in Cross-Domain Few-Shot Learning~\cite{fu2021meta,sun2021explanation,fu2023styleadv}, we evaluate our method on four challenging target domains: CropDiseases~\cite{34699834fa624a3bbc2fae48eb151339}, EuroSAT~\cite{helber2019eurosat}, ISIC2018~\cite{codella2019skin}, and ChestX~\cite{Wang_2017}. These datasets span diverse domains (agriculture, remote sensing, and medical imaging), presenting significant domain shifts from general-source datasets.

\vspace{0.1cm}
\noindent\textbf{Implementation Details.} We adopt the CLIP (ViT-B/16) model as our backbone and perform direct fine-tuning on the target domains without using any source data. We set k = 0.3 and temper ratio = 0.5, and we apply our method in layer 8 and layer 10. Models are trained for 100 epochs,  Following standard practice~\cite{zanella2024low}, we employ data augmentation on support samples during each training episode.
Evaluation is conducted under the 1-shot and 5-shot settings. We report the mean classification accuracy over 800 repeated trials for 1-shot and 400 trials for 5-shot learning to ensure statistical reliability.
\begin{table}[t]
	\begin{center}
    \caption{Ablation study on the 5-way 5-shot task.}\vspace{-0.3cm}
		\resizebox{0.48\textwidth}{!}{
			\begin{tabular}{lcccccc}
				\toprule
				Inhibit&Enhance & CropDisease & EuroSAT & ISIC2018 & ChestX & Ave. \\
				\midrule
				- & - &96.21 & 92.52 & 51.10& 24.13 & 65.99 \\
				
				  $\checkmark$ & -& 97.21& 93.26& 55.00 & 25.15 &  67.66\\
				
				 -&$\checkmark$ &97.24  & 92.99 & 53.72& 24.98 & 67.23  \\
                 
				 $\checkmark$&$\checkmark$ & \textbf{97.42} & \textbf{93.49} & \textbf{56.73} & \textbf{26.12} & \textbf{68.44} \\
				\bottomrule
		\end{tabular}}\vspace{-0.5cm}
    \label{tab:ablation}
    \end{center}
\end{table}
\subsection{Comparison with State-of-the-Art Works}
We conduct comprehensive comparisons with a wide range of representative CDFSL methods~\cite{fu2023styleadv,zou2024flatten,ma2024reconstruction,zoucloser,zouattention,yi2025revisiting,yi2025random,yazdanpanah2022visual,xu2024enhancing,xu2024step} under various experimental settings, including different backbones, the use of source domain data, and whether fine-tuning is applied on the target domain. As shown in Tab.~\ref{tab:sotas}, our method consistently achieves top average performance across both 1-shot and 5-shot settings, outperforming all current state-of-the-art works.  These results validate that our method not only enhances discriminative learning but also generalizes robustly across diverse target domains and fine-tuning scenarios. 
\vspace{-0.1cm}
\subsection{Ablation Study}
\vspace{-0.1cm}
We conduct ablation experiments to validate the contribution of each module in our method, which consists of two core components: one that suppresses sink tokens and another that enhances discriminative token learning. Using CLIP-LoRA-Vision as the baseline, we progressively integrate each module and evaluate under the 5-way 5-shot setting across four CDFSL benchmarks. As summarized in Tab.~\ref{tab:ablation}, both modules individually improve performance, with the suppression mechanism effectively reducing interference from trivial patterns and the enhancement module promoting semantically discriminative learning. When combined, the full model achieves the highest accuracy across all datasets, demonstrating that jointly inhibiting sink tokens and reinforcing discriminative features leads to more robust cross-modal classification performance in cross-domain few-shot scenarios.

\begin{figure}[t]
	\centering
	\includegraphics[width=0.95\columnwidth]{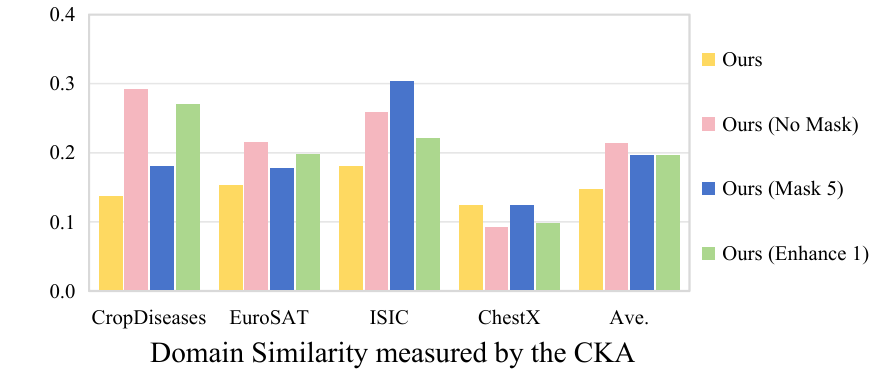}
    \vspace{-0.3cm}
	\caption{ Comparison of CKA similarities across four testing scenarios, showing method shifts the learning of sink tokens to discriminative tokens. Detailed analysis reveals that after applying our method, ``Sum=5'' (sink) tokens no longer primarily carry domain information, while ``Sum=1'' (discriminative) tokens begin to absorb more domain-specific characteristics.}\vspace{-0.2cm}
	\label{fig:final_cka}
\end{figure}

\begin{figure}[t]
	\centering
	\includegraphics[width=1.0\columnwidth]{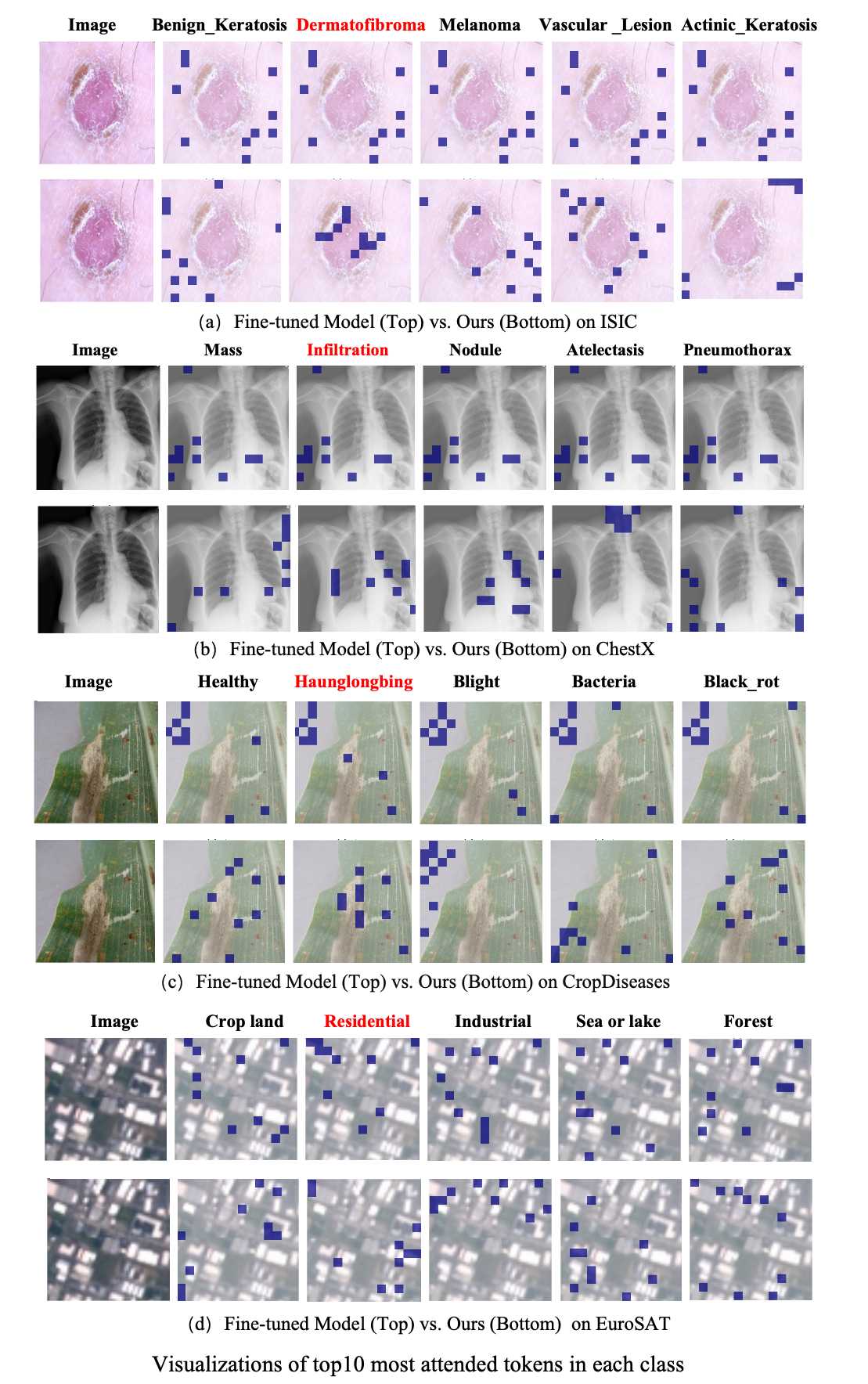}
    \vspace{-0.7cm}
	\caption{ Standard fine-tuned models consistently show identical attended tokens, while our models can attend to different discriminative tokens in the image for different classes.}\vspace{-0.3cm}
	\label{fig:more_visual}
\end{figure}

\subsection{Shifting the learning from sink tokens to discriminative tokens}
\subsubsection{Quantitative Study}
We conduct Centered Kernel Alignment (CKA) analysis to validate that our method shifts the learning from sink tokens to discriminative tokens in a quantitative way. As shown in Fig.~\ref{fig:final_cka}, we compare four experimental settings: after fine-tuning by our methods, in the evaluation phase: (1) applying our method for testing; (2) using standard testing; (3) masking ``Sum=5'' sink tokens during testing; and (4) enhancing ``Sum=1'' discriminative tokens during testing.

The results reveal a significant redistribution of domain information roles: when masking ``Sum=5'' sink tokens after fine-tuning with our method, the decreased CKA similarity indicates increased domain-specific information, demonstrating that these original sink tokens no longer serve as primary carriers of domain information. Conversely, when enhancing ``Sum=1'' discriminative tokens, the reduced CKA similarity reveals that these tokens begin to absorb more domain information, confirming our method shifts the learning from sink tokens to discriminative tokens.
\vspace{-0.1cm}
\subsubsection{Qualitative Study}
\vspace{-0.1cm}
We provide visualization results comparing attention patterns across different settings in Fig.~\ref{fig:more_visual}. 
We can see that the standard fine-tuned model consistently shows identical attended tokens, indicating a severe attention sink problem. In contrast, our model learns to focus on class-specific discriminative regions while accurately identifying image areas most relevant to the correct categories, resulting in more precise and semantically meaningful attention alignment that contributes to improved classification performance.

\begin{figure}[t]
	\centering
	\includegraphics[width=1.0\columnwidth]{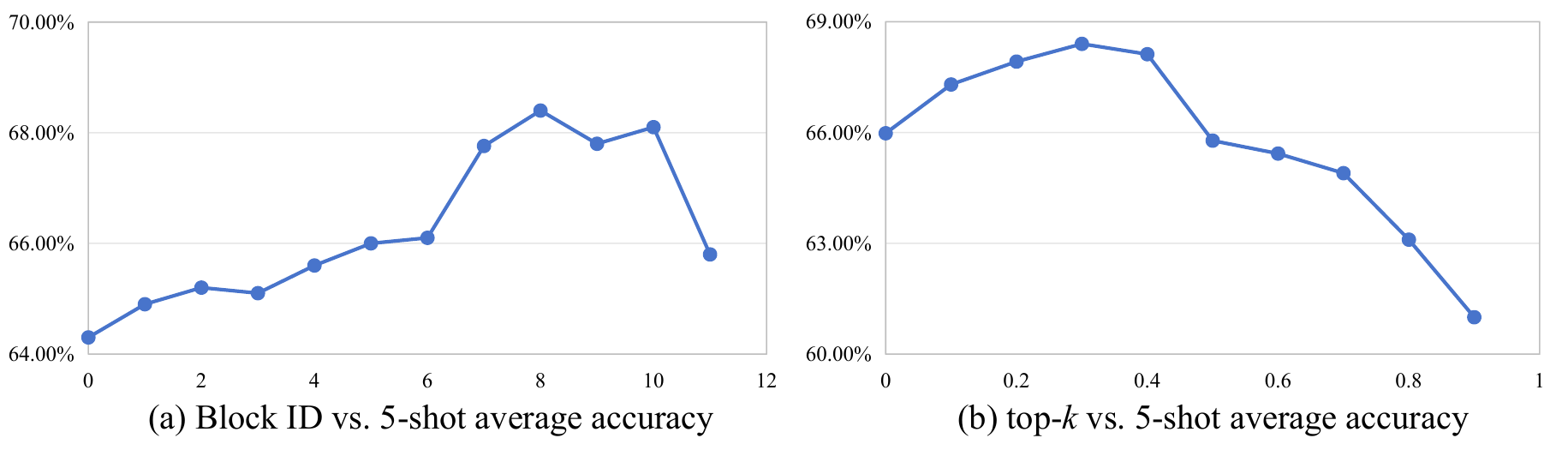}
    \vspace{-0.6cm}
	\caption{ (a) The results show that inserting our method after deeper blocks (particularly after the 8th block) yields optimal performance, validating our analysis that semantic discrimination and attention sink predominantly occur in deeper network stages where complex patterns emerge. (b) Performance variation with different top-k ratios, showing optimal results at 0.3 and performance degradation beyond 0.5, confirming that focusing on the top 30\% most class-sensitive tokens provides the best balance between capturing discriminative patterns and avoiding non-discriminative noise.}\vspace{-0.3cm}
	\label{fig:hyper}
\end{figure}
\vspace{-0.1cm}
\subsubsection{Sensitivity Study of Hyper-parameters}
\vspace{-0.1cm}
We conduct a comprehensive analysis of the hyperparameters in our method as presented in Fig.~\ref{fig:hyper}. The results demonstrate that our approach achieves optimal performance with properly configured parameters while providing valuable insights into the model's behavior.

(1) \textbf{Insertion Position of Our Method.} As shown in Fig.~\ref{fig:hyper}a, we evaluate the performance impact when inserting our method at different transformer blocks. The results reveal that applying our method in deep blocks (after 7th block onward) consistently improves performance, with the optimal position being after the 8th block. This finding aligns with our previous analysis that shallow layers primarily capture low-level features with weak semantic awareness, while deeper layers develop more complex semantic patterns. The attention sink phenomenon becomes most pronounced in these deeper layers where complex patterns emerge, making our intervention most effective at this stage. Applying our method too early (before 4th block) shows limited benefits, confirming that semantic discrimination primarily occurs in deeper network stages.

(2) \textbf{Selection of Top-k Ratio.} Fig.~\ref{fig:hyper}b illustrates the sensitivity of performance to the top-k ratio for identifying sink and discriminative tokens. The results show that performance improves as the ratio increases from 0.1 to 0.3, plateaus between 0.3-0.4, and begins to degrade beyond 0.5. The optimal ratio of 0.3 indicates that focusing on the top 30\% most class-sensitive tokens provides the best balance between capturing hard patterns and avoiding interference from sink tokens. The performance degradation at higher ratios (e.g., 0.5) suggests that including excessive tokens introduces noise from other patterns, exacerbating the attention sink problem. This validates our design choice to selectively target the most critical tokens rather than employing a broad-brush approach.
\vspace{-0.1cm}
\section{Related Work}
\vspace{-0.1cm}

\subsection{Cross-Domain Few-Shot Learning (CDFSL)}
Cross-Domain Few-Shot Learning (CDFSL) aims to develop models capable of rapid generalization to novel target domains with only a limited number of examples\cite{zhang2024learning,zhang2024micm}. Existing methodologies are broadly categorized into two streams: meta-learning-based approaches\cite{fu2022wavesan,hu2022adversarial}, which train models through episodic tasks to quickly adapt to new feature distributions, and transfer learning-based methods\cite{Zhou_2023_CVPR,yi2025revisiting,yi2025random}, which focus on enhancing the model's inherent generalization ability during the source training phase. A more challenging and practical setting, Source-Free CDFSL (SF-CDFSL)\cite{zhang2026reclaiming,zhang2026mind}, has recently emerged, where the source domain data is entirely inaccessible during adaptation, placing full emphasis on the fine-tuning strategy for the target domain. While prior works like AttnTemp\cite{zouattention} have explored mitigating attention bias, they primarily do so during direct transfer or source-domain training. In contrast, the influence of discriminative learning and the specific challenge of visual attention sinks within CLIP-based models have not been explored.

\subsection{Attention Sink}
Attention sink, wherein tokens with minimal semantic content (e.g., punctuation, BOS, or background patches) consistently attract disproportionately high attention scores, has been recognized as a fundamental characteristic of transformer-based architectures across both language and vision domains\cite{xiao2023efficient}. In language models, this behavior is often attributed to the softmax constraint in attention mechanisms, leading to stable but semantically sparse attention distributions\cite{gu2024attention}. Recent studies have extended this concept to the visual domain, identifying "visual attention sinks" in Large Multimodal Models where certain image patches (e.g., uniform backgrounds) absorb significant attention without contributing meaningfully to downstream tasks\cite{kang2025see}. While several methods\cite{gu2024attention,jung2025visual,arif2025paint,darcet2024vision} have been proposed to mitigate or exploit attention sinks in unimodal language or vision transformer, such as attention recalibration or token removal strategies, the role and impact in multimodal fine-tuning scenarios, particularly under cross-domain few-shot settings, remain largely unexamined. Our work bridges this gap by systematically investigating how visual attention sinks manifest and are amplified during few-shot adaptation of vision-language models, and proposes a targeted approach to suppress their interference while enhancing discriminative cross-modal alignment.

\vspace{-0.1cm}
\section{Conclusion}
\vspace{-0.1cm}
We find a phenomenon that few-shot fine-tuning in CDFSL unexpectedly exacerbates the visual attention sink. We delve into it for an interpretation, finding that the shortcut of domain adaptation is the cause. Based on it, we propose a method to explicitly suppress the model's reliance on shortcut learning. Experiments validate our rationale and effectiveness, showing new state-of-the-art performance.

\section*{Acknowledgments}

This work is supported by the National Natural Science Foundation of China under grants 62206102; the National Key Research and Development Program of China under grant 2024YFC3307900; the National Natural Science Foundation of China under grants 62436003, 62376103 and 62302184; Major Science and Technology Project of Hubei Province under grant 2025BAB011 and 2024BAA008; Hubei Science and Technology Talent Service Project under grant 2024DJC078; and Ant Group through CCF-Ant Research Fund. The computation is completed in the HPC Platform of Huazhong University of Science and Technology.


{    \small
    \bibliographystyle{ieeenat_fullname}
    \bibliography{main}
}
\clearpage
\clearpage
\setcounter{page}{1}
\maketitlesupplementary

\section{Detailed Dataset Description}
\label{sec:Detailed Dataset Description}

\begin{figure}[h]
\vspace{-0.5cm}
	\centering	\includegraphics[width=1.0\columnwidth]{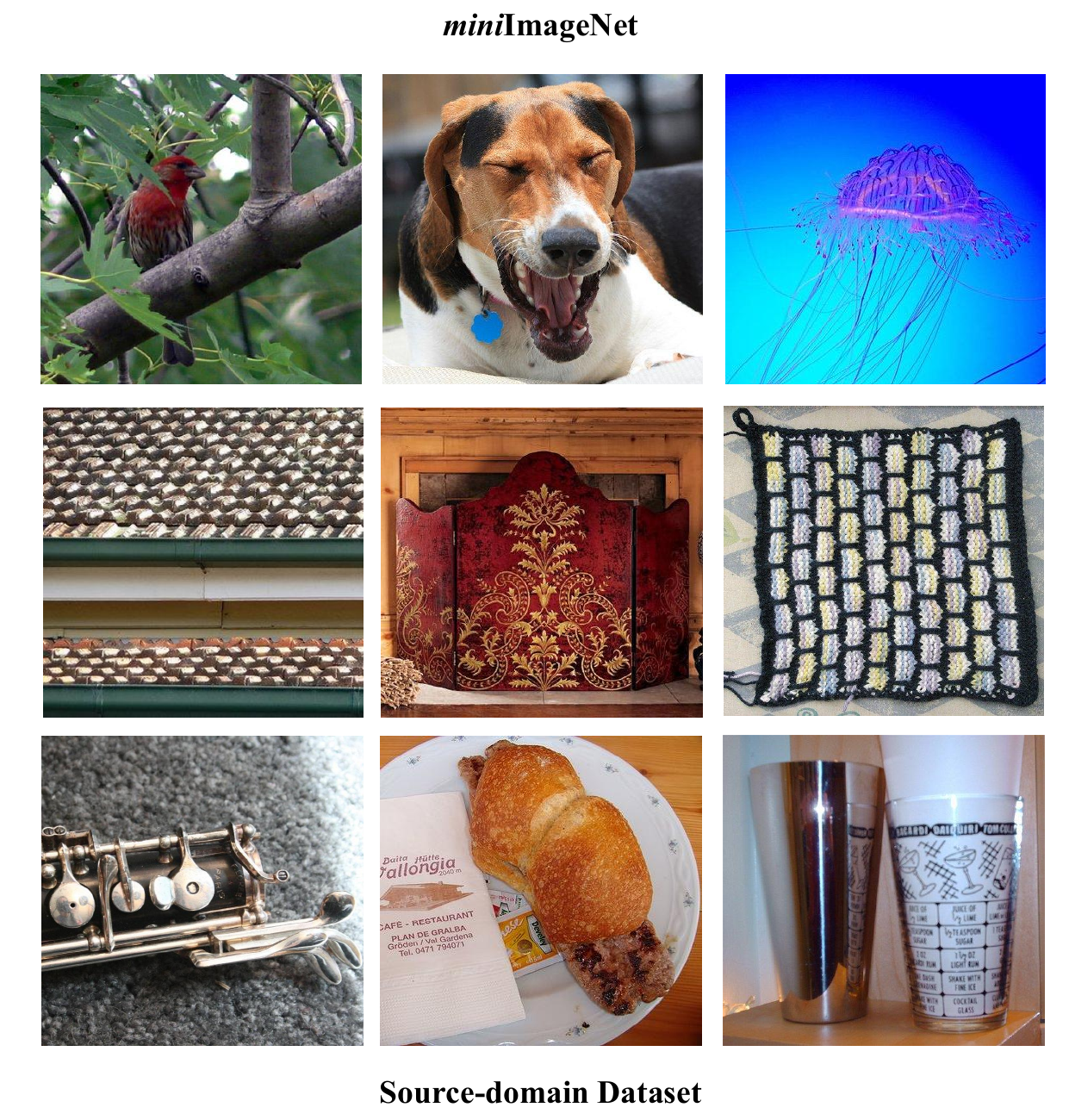}
    \vspace{-0.5cm}
	\caption{Representative samples from the source-domain \textit{mini}ImageNet dataset.}
	\label{fig:source}
\end{figure}

\textbf{\textit{mini}ImageNet}~\cite{Vinyals2016Matching} is a widely adopted benchmark in meta-learning and few-shot learning, comprising a curated subset of the original ImageNet~\cite{deng2009imagenet} dataset. The dataset contains 60,000 color images distributed across 100 object categories, with each category consisting of 600 samples of size $84\times84$ pixels. As illustrated in Fig.~\ref{fig:source}, \textit{mini}ImageNet encompasses diverse real-world scenes with varying contextual elements, including both human-centric scenarios and natural environments. 

\textbf{CropDiseases}~\cite{34699834fa624a3bbc2fae48eb151339} provides a specialized agricultural dataset for plant disease recognition, containing 43,456 high-resolution images across 38 disease categories. The dataset exhibits huge domain shift from natural images, featuring detailed close-ups of infected and healthy plant specimens with high intra-class similarity.

\textbf{EuroSAT}~\cite{helber2019eurosat} offers a remote sensing benchmark for land use classification, comprising 27,000 satellite images categorized into 10 distinct land cover types. The aerial perspective and absence of conventional photographic distortions create substantial domain gap from natural image datasets.

\begin{figure}[h]
	\centering
   \includegraphics[width=1.0\columnwidth]{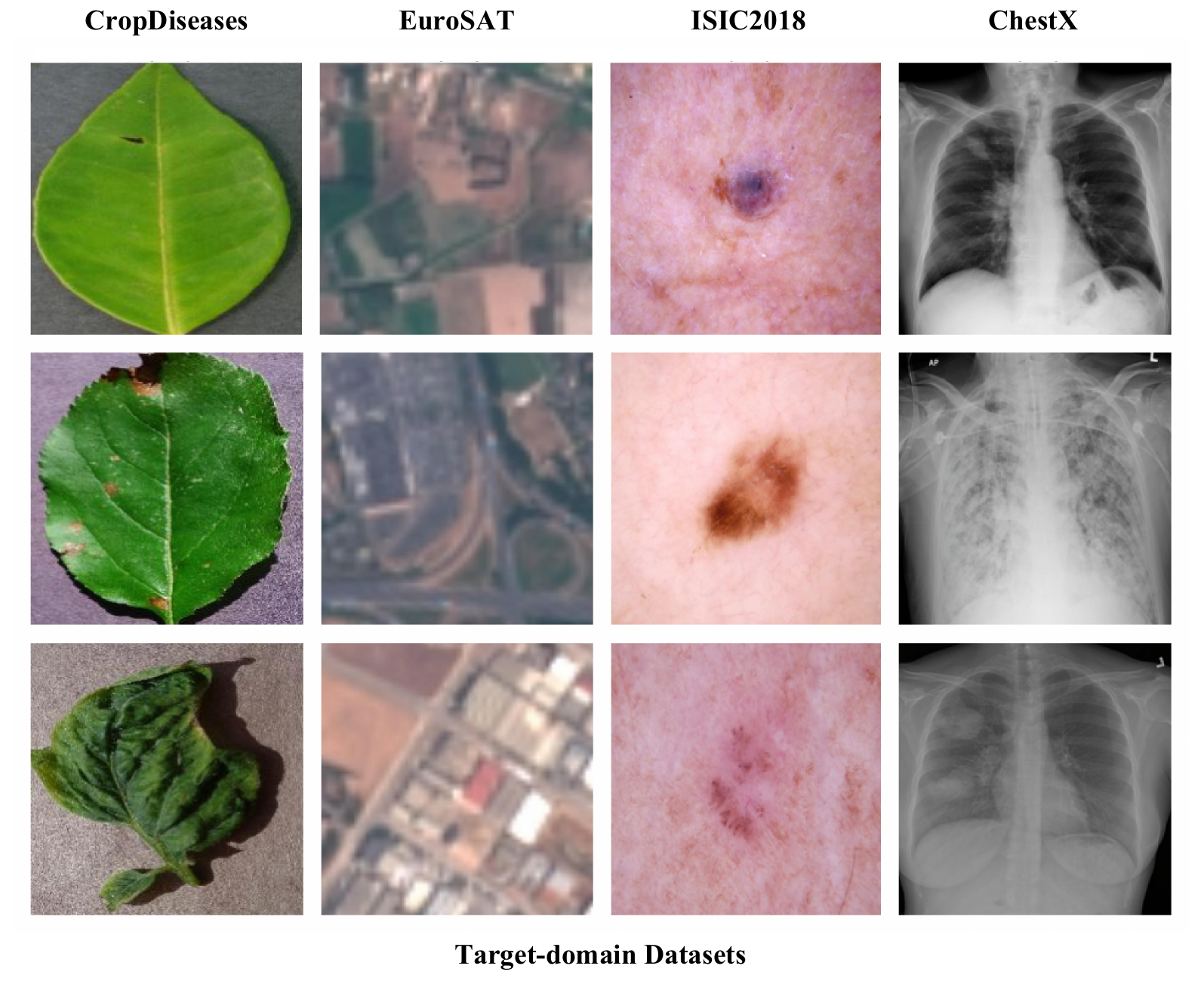}
    \vspace{-0.4cm}
	\caption{Sample images from the four target-domain benchmarks: CropDiseases (agricultural), EuroSAT (remote sensing), ISIC2018 (dermatological), and ChestX (medical imaging).}
	\label{fig:target}
\end{figure}

\textbf{ISIC2018}~\cite{codella2019skin} constitutes a medical imaging dataset for dermatological analysis, containing 10,015 dermoscopic images across 7 skin lesion categories. The clinical nature and specialized visual patterns represent huge domain shift from conventional natural images.

\textbf{ChestX}~\cite{Wang_2017} provides a medical radiology dataset of 25,847 chest X-ray images distributed across 7 thoracic conditions. This dataset exhibits the most substantial domain gap due to its monochromatic medical imaging modality, anatomical focus, and absence of natural scene characteristics.

As depicted in Fig.~\ref{fig:target}, these four target domains: agricultural, remote sensing, dermatological, and medical radiology, which collectively represent challenging cross-domain scenarios with progressively increasing domain shifts from the source domain.

\section{Detailed Descriptions of the CKA}

Following established practices in domain similarity measurement~\cite{oh2022understanding,kornblith2019similarity}, we employ Centered Kernel Alignment (CKA) to quantitatively assess the similarity between feature representations across different domains. CKA is a robust statistical method specifically designed to compare high-dimensional representations learned by neural networks, with particular strength in analyzing cross-domain feature relationships.

The CKA methodology operates through the following computational process. Given two sets of feature representations $X \in \mathbb{R}^{n \times d}$ and $Y \in \mathbb{R}^{n \times d}$ extracted from different domains, we first compute their Gram matrices:
\begin{equation}
K = XX^\top, \quad L = YY^\top
\end{equation}

These matrices capture the inner product relationships between all pairs of samples within their respective feature spaces. To eliminate the influence of data means and ensure proper alignment, we center the Gram matrices using:
\begin{align}
K_c &= HKH \\
L_c &= HLH
\end{align}
where $H = I_n - \frac{1}{n}\mathbf{1}_n\mathbf{1}_n^\top$ denotes the centering matrix, $I_n$ is the identity matrix, and $\mathbf{1}_n$ represents a vector of ones. The final CKA similarity is then computed as:
\begin{equation}
\text{CKA}(X,Y) = \frac{\text{vec}(K_c) \cdot \text{vec}(L_c)}{\|\text{vec}(K_c)\| \|\text{vec}(L_c)\|} = \frac{\text{Tr}(K_c L_c)}{\sqrt{\text{Tr}(K_c^2)\text{Tr}(L_c^2)}}
\end{equation}

This normalized metric quantifies the similarity between the relational structures encoded in the two feature sets, with values ranging from 0 (completely dissimilar) to 1 (identical relationships).

In our work, we leverage CKA to measure domain distance between source and target datasets following~\cite{davari2022reliability}. Given a pre-trained backbone network, we extract features from image batches sampled from different domains and compute CKA similarity after aligning the channel dimensions. The interpretation follows established conventions~\cite{kim2023stability}: higher CKA values indicate greater domain similarity and consequently less domain-specific information in the representations, while lower values suggest stronger domain shift and more domain-characteristic features. This approach provides crucial insights into the model's generalization capabilities and domain adaptation characteristics across diverse data distributions.

\begin{figure*}[!p]
	\centering
\includegraphics[width=1.7\columnwidth]{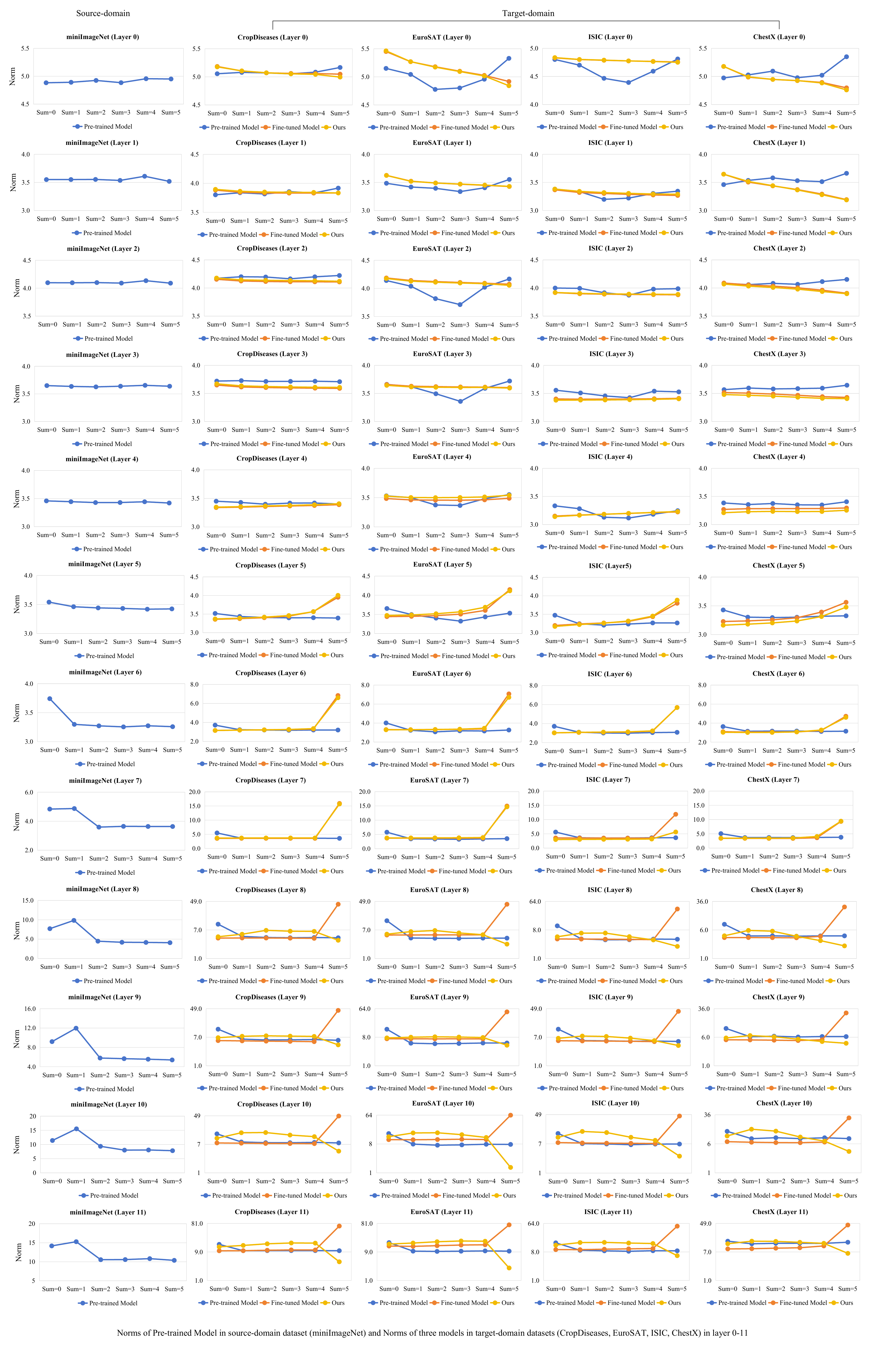}
    \vspace{-0.8cm}
	\caption{At shallow layers, both source and target domain models show similar attention distributions with weak semantic awareness. At deep layers, the source domain model develops specialized attention to discriminative tokens (“Sum=1”), while the target domain fine-tuned model shifts toward non-discriminative tokens (“Sum=5”) with higher norms, demonstrating that fine-tuning, instead of merely cross-domain transfer, serves as the primary driver of attention sink exacerbation.}\vspace{-0.4cm}
	\label{fig:all_layer_norm}
\end{figure*}
\section{More Experiments}

\subsection{Norm distribution of different sum numbers from layer 0 to layer 11}
We provide comprehensive layer-wise analyses of token norm distributions across all transformer layers (0 to 11) in Fig.\ref{fig:all_layer_norm}. The complete visualization across all layers offers deeper insights into the evolution of semantic awareness and attention patterns throughout the network architecture.

As systematically illustrated in the Fig.\ref{fig:all_layer_norm}, several key patterns emerge across the layer hierarchy:

(1) \textbf{Early Layers (0-4)}: Both source and target domains exhibit similar norm distributions regardless of sum scores, confirming that shallow layers primarily process low-level features with minimal semantic discrimination. The uniform distribution across all token types demonstrates the model's initial lack of class-specific awareness in these foundational layers.

(2) \textbf{Middle Layers (5-7)}: A gradual divergence begins to emerge between source and target domains. In the source domain, we observe the initial development of differential norm patterns, with "Sum=1" tokens starting to receive slightly elevated attention. Conversely, target domains maintain relatively flat distributions, indicating stalled development of discriminative patterns due to domain shift.

(3) \textbf{Deep Layers (8-11)}: The critical divergence becomes most pronounced in these final layers. Source domain models exhibit strong specialization, with "Sum=1" discriminative tokens receiving substantially higher norms than other token types. This pattern reflects successful development of class-specific attention mechanisms. In contrast, target domain models show two distinct trajectories: pre-trained models maintain weak discrimination, while fine-tuned models develop inverted patterns with "Sum=5" sink tokens dominating the norm distribution.

The complete layer progression provides compelling evidence for our core thesis: the attention sink exacerbation stems from fine-tuning's impact on deep layer transformations rather than cross-domain transfer alone. While domain shift prevents the natural emergence of discriminative patterns, it is the fine-tuning process that actively drives the model toward non-discriminative sink tokens as an adaptation strategy.

These comprehensive results across all layers and datasets reinforce our interpretation that standard fine-tuning in CDFSL scenarios creates a problematic shortcut, where models prioritize easily alignable but non-discriminative patterns over semantically meaningful features, ultimately limiting their generalization capability in target domains.

\begin{table}[t]
    \vspace{-0.1cm}
	\begin{center}
    \caption{TIR with more backbones.}\vspace{-0.3cm}
		\resizebox{0.48\textwidth}{!}{
			\begin{tabular}{lccccc}
				\toprule
				Method (Shot) & CropDisease & EuroSAT & ISIC2018 & ChestX & Ave. \\
				\midrule
				ViT/SigLIP2 (1-shot) &80.98 & 66.52 & 29.04& 21.34 & 49.47 \\
			
                \textbf{+Ours} & \textbf{84.82} & \textbf{69.87} & \textbf{32.20} & \textbf{21.53} & \textbf{52.11} \\
                 \midrule
                 ViT/PE-Core (1-shot) &87.16 &  78.43 & 39.39& 21.93 & 56.73 \\
			
                \textbf{+Ours} & \textbf{88.84} & \textbf{80.17} & \textbf{40.31} & \textbf{22.23} & \textbf{57.89} \\
                
                 \midrule
                 ViT/SigLIP2 (5-shot) &95.43 & 85.17 & 44.26& 23.68 & 62.14 \\
			
                 \textbf{+Ours} & \textbf{97.05} & \textbf{88.83} & \textbf{51.10} & \textbf{24.61} & \textbf{65.40} \\
                 \midrule
                 ViT/PE-Core (5-shot) &96.97 & 90.46 & 55.22& 25.52 & 67.04 \\
                \textbf{+Ours} & \textbf{97.21} & \textbf{92.60} & \textbf{57.34} & \textbf{25.89} & \textbf{68.26} \\
                 
				\bottomrule
		\end{tabular}}\vspace{-0.4cm}
    \label{tab:more_backbone}
    \end{center}
\end{table}
\begin{table}[h]
    \vspace{-0.35cm}
	\begin{center}
    \caption{TIR with more adaptation strategies.}\vspace{-0.3cm}
		\resizebox{0.48\textwidth}{!}{
			\begin{tabular}{lccccc}
				\toprule
				Method (Shot) & CropDisease & EuroSAT & ISIC2018 & ChestX & Ave. \\
				\midrule
				CLIP + CoOp  &91.85 & 83.27 & 43.31& 22.67 & 60.28 \\
			
                 \textbf{+Ours} & \textbf{92.73} & \textbf{85.33} & \textbf{44.49} & \textbf{24.44} & \textbf{61.75} \\
                 \midrule
                 CLIP + LoRA(Text)  &92.74 & 82.98 & 42.97& 22.84 & 60.38 \\
			
                 \textbf{+Ours} & \textbf{93.19} & \textbf{83.26} & \textbf{47.35} & \textbf{23.33} & \textbf{61.78} \\
                 \midrule
                 CLIP + MaPLe &96.20 & 90.00 & 50.47& 24.11 & 65.20 \\
			
                 \textbf{+Ours} & \textbf{96.80} & \textbf{93.35} & \textbf{54.04} & \textbf{25.75} & \textbf{ 67.49} \\
                  \midrule
                 CLIP + LoRA(Vision)  &96.21 & 92.52 & 51.10& 24.13 & 65.99 \\ 
			
                 \textbf{+Ours} & \textbf{97.42} & \textbf{93.49} & \textbf{56.73} & \textbf{26.12} & \textbf{68.44} \\
				\bottomrule
		\end{tabular}}\vspace{-0.4cm}
    \label{tab:more_strategies}
    \end{center}
\end{table}
\subsection{Extending to more backbones}
We evaluate our method on two representative backbones, SigLIP2~\cite{tschannen2025siglip} and PE-Core~\cite{bolya2025perception}, with results reported in Tab. \ref{tab:more_backbone}. Our approach consistently achieves improvements over the baselines in both the 1-shot and 5-shot settings, demonstrating its effectiveness across different architectures and few-shot scenarios.

\subsection{Experiments with other finetuning strategies}
We implement TIR with more fine-tuning strategies in Tab.~\ref{tab:more_strategies}, and observe consistent improvements across all settings. Specifically, when applied to CoOp~\cite{zhou2022conditional}, LoRA~\cite{hu2022lora} (Text), MaPLe~\cite{khattak2023maple}, and LoRA~\cite{hu2022lora}  (Vision), our method yields higher accuracy on each individual dataset. Notably, the gains are particularly pronounced on the more challenging ISIC2018 and ChestX datasets, and the overall average improvement reaches up to 2.45\% when combined with LoRA (Vision). These results demonstrate that TIR generalizes effectively across diverse adaptation strategies, consistently boosting few-shot performance.
\begin{table}[t]
	\begin{center}
    \caption{Comparisons with three types of attention sink methods.}\vspace{-0.3cm}
		\resizebox{0.48\textwidth}{!}{
			\begin{tabular}{lcccccc}
				\toprule
				Method & Mark & CropDisease & EuroSAT & ISIC2018 & ChestX & Ave. \\
				\midrule
				Baseline & - &96.21 & 92.52 & 51.10& 24.13 & 65.99 \\
				
				  SPARC & ICML25 & 70.60& 62.61& 40.58 & 22.70 &  49.12\\
				
				  Registers & NeurIPS25 &96.54  & 90.67 & 50.53& 23.56 & 65.33  \\
                 
				  PAINT & CVPRW25 & 96.88 & 91.65 & 50.58 & 23.62 & 65.68 \\
                 \textbf{TIR} & Ours & \textbf{97.42} & \textbf{93.49} & \textbf{56.73} & \textbf{26.12} & \textbf{68.44} \\
				\bottomrule
		\end{tabular}}\vspace{-0.4cm}
    \label{tab:attnsink_method}
    \end{center}
\end{table}

\subsection{Comparison with other attention sink works}

We compare our method with existing approaches designed to mitigate attention sink phenomena, including SPARC~\cite{jung2025visual}, Registers~\cite{darcet2024vision}, and PAINT~\cite{arif2025paint}. As shown in Tab.~\ref{tab:attnsink_method}, these methods either underperform or yield only marginal improvements over the baseline in the source-free cross-domain few-shot learning setting. Specifically, SPARC causes a substantial performance drop across all datasets, achieving an average accuracy of only 49.12\%. Registers and PAINT obtain averages of 65.33\% and 65.68\%, respectively. Both are slightly below the baseline's 65.99\%. In contrast, our method consistently improves upon the baseline across every dataset, attaining the highest average accuracy of 68.44\%. These results demonstrate the effectiveness of our approach in addressing attention sink challenges under domain shift and few-shot constraints. 

\begin{table}[t]
	\begin{center}
    \vspace{-0.3cm}
    \caption{Comprehensive ablation study of different design choices for token recalibration on the 5-way 5-shot task.}
		\vspace{-0.2cm}
		\resizebox{0.48\textwidth}{!}{
			\begin{tabular}{lcccccc}
				\toprule
				Method & CropDisease & EuroSAT & ISIC2018 & ChestX & Ave. \\
				\midrule
				Baseline&96.21 & 92.52 & 51.10& 24.13 & 65.99 \\
				
				  Only ``Sum=5'' weight=0& 97.21& 93.26& 55.00 & 25.15 &  67.66\\
				
				 Only ``Sum=1'' weight=2&97.24  & 92.99 & 53.72& 24.98 & 67.23  \\
                 
				 \textbf{Ours} & 97.42& \textbf{93.49} & \textbf{56.73} & \textbf{26.12} & \textbf{68.44} \\
                 \midrule
                 ``Sum=5'' weight=0 and ``Sum=1'' weight=2& \textbf{97.43} & 93.40& 56.72 & 25.92 &  68.37\\
                 \midrule
                 Only ``Sum=5'' weight=0.5& 96.22&92.54& 52.42 & 24.32 &  66.38\\
                 Only ``Sum=5'' weight=0.1& 96.40&92.60& 52.75 & 24.61 &  66.59\\
                 \midrule
                 Only ``Sum=1'' weight=1.5& 97.18&92.96& 53.47 & 24.87 &  67.12\\
                 Only ``Sum=1'' weight=3& 97.22&92.83& 53.44 & 24.81 &  67.08\\
                 Only ``Sum=1'' weight=4& 97.15&92.79& 53.43 & 24.96 &  67.08\\
                 Only ``Sum=1'' weight=5& 97.19&92.82& 53.37 & 24.65 &  67.01\\
				\bottomrule
		\end{tabular}}\vspace{-0.4cm}
    \label{tab:design_choice}
    \end{center}
\end{table}
\subsection{Other design choices}
We present a comprehensive ablation study examining various design choices for our token recalibration strategy. The experimental results, summarized in the Tab.\ref{tab:design_choice}, provide valuable insights into the optimal configuration for addressing attention sink in SF-CDFSL scenarios.

(1) \textbf{Critical Role of Sum=5 Suppression}: The experimental results demonstrate that completely suppressing Sum=5 tokens (setting their weights to 0) yields the most significant performance improvements. Alternative suppression strengths (weights of 0.5 or 0.1) prove substantially less effective, indicating that strong suppression of these non-discriminative sink tokens is essential for mitigating attention sink exacerbation.

(2) \textbf{Robustness of Sum=1 Enhancement}: Interestingly, the enhancement of Sum=1 discriminative tokens shows remarkable robustness to specific weight values. Our experiments with enhancement weights of 1.5, 3, 4, and 5 reveal that any value greater than 1 produces comparable performance gains. This insensitivity to the exact enhancement magnitude suggests that the crucial factor is simply providing positive reinforcement to class-specific tokens rather than fine-tuned weight optimization.

(3) \textbf{Effectiveness of Minimal Intervention}: Perhaps most notably, the combination of solely suppressing Sum=5 tokens while enhancing Sum=1 tokens achieves performance nearly equivalent to our complete method (which additionally handles intermediate Sum=2,3,4 tokens). This finding indicates that the core attention sink problem in SF-CDFSL primarily stems from the extreme cases of completely non-discriminative and highly discriminative tokens.

These insights lead to a simplified and generalizable approach for SF-CDFSL: in K-way N-shot settings, setting the weight of Sum=K tokens to 0 while assigning Sum=1 tokens any weight greater than 1 effectively addresses the attention sink exacerbation problem. This simplified strategy maintains the performance benefits of our complete method while offering greater practicality and easier implementation for various few-shot learning configurations.

The robustness of this approach across different enhancement weights and its effectiveness with minimal intervention make it particularly suitable for real-world SF-CDFSL applications where architectural simplicity and parameter insensitivity are valuable attributes.


\end{document}